\pdfoutput=1

\documentclass[11pt]{article}

\usepackage[final]{acl}

\usepackage{times}
\usepackage{latexsym}

\usepackage[T1]{fontenc}

\usepackage[utf8]{inputenc}

\usepackage{microtype}

\usepackage{inconsolata}

\usepackage{graphicx}

\usepackage{multirow}

\title{FastDraft: How to Train Your Draft}

\author{Ofir Zafrir\textsuperscript{*}, 
Igor Margulis\textsuperscript{*},
Dorin Shteyman\textsuperscript{*},
Shira Guskin,
Guy Boudoukh\\
Intel Labs\\
\texttt{\{ofir.zafrir,igor.margulis,dorin.shteyman\}@intel.com}\\
\texttt{\{shira.guskin,guy.boudoukh\}@intel.com}}

\begin{document}
\maketitle
\def\thefootnote{*}\footnotetext{Equal contribution}\def\thefootnote{\arabic{footnote}
}
\begin{abstract}
Speculative Decoding has gained popularity as an effective technique for accelerating the auto-regressive inference process of Large Language Models. 
However, Speculative Decoding entirely relies on the availability of efficient draft models, which are often lacking for many existing language models due to a stringent constraint of vocabulary compatibility.
In this work we introduce FastDraft, a novel and efficient approach for pre-training and aligning a draft model to any large language model by incorporating efficient pre-training, followed by fine-tuning over synthetic datasets generated by the target model.
We demonstrate FastDraft by training two highly parameter efficient drafts for the popular Phi-3-mini and Llama-3.1-8B models.
Using FastDraft, we were able to produce a draft model with approximately 10 billion tokens on a single server with 8 Intel\textsuperscript{\tiny\textregistered} Gaudi\textsuperscript{\tiny\textregistered} 2 accelerators in under 24 hours.
Our results show that the draft model achieves impressive results in key metrics of acceptance rate, block efficiency and up to 3x memory bound speed up when evaluated on code completion and up to 2x in summarization, text completion and instruction tasks.
We validate our theoretical findings through benchmarking on the latest Intel\textsuperscript{\tiny\textregistered}  Core\textsuperscript{\tiny\texttrademark} Ultra, achieving a wall-clock time speedup of up to 2x, indicating a significant reduction in runtime.
Due to its high quality, FastDraft unlocks large language models inference on AI-PC and other edge-devices.
\end{abstract}

\section{Introduction}
The advent of Transformer architectures has fundamentally reshaped the field of natural language processing (NLP).
In recent years, Transformer-based models have achieved remarkable success across a broad spectrum of natural language understanding and generation tasks~\citep{achiam2023gpt,team2023gemini}.
Their exceptional performance, particularly in large language models (LLMs), has made them highly desirable for deployment in numerous applications, ranging from conversational systems to content generation and beyond.
Despite their outstanding performance, LLMs suffer from slow inference speed due to substantial memory bandwidth requirements and the sequential nature of auto-regressive generation (ARG).
The introduction of Speculative Decoding (SD) \citep{leviathan2023fast} offers a promising solution for accelerating ARG without sacrificing generation quality, making it a compelling approach for improving LLM inference efficiency.
SD utilizes a draft language model (LM) to generate a sequence of tokens auto-regressively, while the target model validates the batched tokens in parallel. 
In certain applications, SD can achieve a 2-3x speedup in LLM inference without compromising the generation quality of the target model.
Achieving significant speedup with SD requires a high-quality draft model that is both efficient and well-aligned with the target.
To date, such draft models remain scarce, even for widely used open-source LLMs~\citep{dubey2024llama, abdin2024phi} due to vocabulary incompatibility.
To address this limitation, we propose FastDraft, a method for producing hardware (HW) efficient draft models that are orders of magnitude smaller than their corresponding target models.

While extensive research has focused on training and data generation for high-quality LLMs \citep{kaplan2020scaling, longpre2023pretrainer}, these frameworks are not necessarily applicable to training draft models for SD.
LLMs are typically trained to generate helpful, high-quality responses, whereas draft models should be trained to generate sequences that are likely to be accepted by the target model.
We explore previously unexamined aspects of draft model training for SD.
Our contributions include:
(1) introducing a method for producing quality and highly efficient draft models with low resource requirements for any given target LLM and demonstrate it by training and benchmarking a draft for Phi-3-mini,
(2) conducting extensive ablation studies on pre-training data size, pre-training for both code and natural language, target-draft alignment via knowledge distillation (KD) and HW-aware draft architectures, and
(3) demonstrating the scalability of FastDraft by training a draft model for Llama-3.1-8B-Instruct, achieving performance improvements comparable to those attained for Phi-3-mini.
We demonstrate significant improvements in key metrics using FastDraft, leading to theoretical speedups of up to 3x as measured by the Memory Bound Speedup (MBSU) metric.
We further validate our theoretical findings by evaluating the wall-clock time speedup achieved with our approach on the latest 
Intel\textsuperscript{\tiny\textregistered} Core\textsuperscript{\tiny\texttrademark} Ultra.
Our results demonstrate an average 1.5x speedup for natural language tasks and an average 2x speedup for code completion tasks.
According to our findings, the small size of the draft model and the limited amount of data required to produce a high-quality draft enabled us to successfully train and align a draft model to Phi-3-mini, end-to-end, in under 24 hours using a single node with 8 Intel\textsuperscript{\tiny\textregistered}  Gaudi\textsuperscript{\tiny\textregistered} 2 accelerators.

\section{Related work}
The widespread adoption of LLMs in cloud and edge devices has driven a significant body of research focused on developing alternative strategies for ARG to address the low inference efficiency of LLMs \citep{santilli2023accelerating, ghazvininejad2019mask, stern2018blockwise}.
However, many of these approaches compromise generation quality or require additional training data and architectural modifications. 
The introduction of speculative decoding as a lossless approach for accelerating LLM inference \citep{leviathan2023fast} has inspired a new wave of follow-up research.
Some studies propose using plug-in prediction heads as a drafting mechanism \citep{zhang2024recurrent, cai2024medusa}, while others focus on improving the serving latency of stand-alone draft models \citep{sun2024triforce, chen2023cascade, miao2023specinfer}.
In contrast, our work focuses on directly enhancing stand-alone draft model's capabilities through pre-training and fine-tuning.
We provide more details on the comparison with these methods in Appendix~\ref{app:eagle-comparison}.
Other works apply KD to draft models to improve alignment with the target model.
These studies explore various divergence functions for the KD algorithm, rather than relying solely on the commonly used Kullback–Leibler Divergence (KLD).
For instance, \cite{zhou2023distillspec} proposed Total Variation Distance (TVD) based on Corollary 3.6 in \citep{leviathan2023fast}, which posits that minimizing TVD maximizes the token-level acceptance rate.
\cite{goel2024direct} further developed this approach with TVD++, fine-tuning their model, pre-trained on 600 billion tokens.
Another study, \cite{yan2024decoding}, focuses on enhancing the HW efficiency of draft models by extensively analyzing the trade-off between time latency and acceptance rates, rather than relying solely on the latter.
While their work emphasizes the approach for obtaining a draft model by pruning of the shelf models, our method introduces an efficient approach for pre-training and aligning a draft model.
\cite{li2024eagle} proposes training a compact draft model based on the target model, using a relatively limited dataset. 
The primary aim of the paper is to develop a draft architecture that effectively utilizes the target model's hidden representations and weights.
In contrast, our paper centers on training and aligning any draft architecture that shares only the vocabulary with the target model.

\section{Speculative decoding}
\label{sec:sd}
SD is a lossless decoding paradigm introduced by \cite{leviathan2023fast} for accelerating ARG with LLMs.
It is inspired by speculative execution \citep{hennessy2017computer} and aims to mitigate the inherent latency bottleneck caused by the sequential nature of ARG~\citep{pope2023efficiently}.
SD employs a draft LM to generate a block of \(\gamma\) candidate tokens.
The LLM, referred to as the target model, then processes these candidate tokens in parallel.
The algorithm examines each token's probability distribution, calculated by both the target and draft models, to determine whether the token should be accepted or rejected.
As a result, any LM can function as a draft model, provided it shares the same vocabulary as the target model. 
However, since an LM's vocabulary is fixed during pre-training, leveraging existing models as draft models is only feasible if they were pre-trained on the same vocabulary.
A key metric for benchmarking ARG inference performance is time per output token (TPOT), which represents the expected latency for generating a single output token, excluding pre-filling latency.\footnote{Pre-filling refers to the action of populating the key-value cache with the input tokens' information.}
In SD, TPOT is a function of the draft model latency \( l_D \), the speculation block size \( \gamma \), the corresponding expected block efficiency\footnote{The expected number of tokens accepted within a speculation block.} \( \tau^{\gamma} \), and the target model latency \( l_T^{\gamma} \).
The equation is expressed as:

\begin{equation}
    \label{eq:sd-tpot}
    \mathrm{TPOT_{SD}} = \frac{l_D \times \gamma + l_T^{\gamma}}{\tau^{\gamma}}
\end{equation}

In comparison, for traditional ARG, TPOT is simply \( \mathrm{TPOT_{AR}} = l^1_T \).
The expected speedup is calculated as follows:

\begin{equation}
    \label{eq:sd-speedup}
    \frac{\mathrm{TPOT_{AR}}}{\mathrm{TPOT_{SD}}} = \left( \frac{l_D}{l_T^1} \times \gamma + \frac{l_T^\gamma}{l_T^1} \right)^{-1}\times\tau^\gamma
\end{equation}


From Equation~\ref{eq:sd-speedup}, two key requirements emerge for SD to yield meaningful speedups.
First, the latency for generating the speculated tokens block must be negligible compared to the target model's latency.
Second, the increase in target model latency for block size \( \gamma \) should be insignificant compared to the latency for block size $1$.
The latter condition generally holds for most popular LLMs with sufficiently small \( \gamma \), while the former depends on the availability of draft models that meet the vocabulary constraint.

\section{FastDraft: Build your own draft model}
\subsection{Draft architecture \& pre-training}
\label{sec:pre-training}
The draft architecture imposes only one strict requirement, it must produce a probability distribution over the target's vocabulary.
Beyond this, the design is flexible.
Nevertheless, certain factors should be considered when selecting the draft's architecture, with latency being the primary concern, as discussed in Section~\ref{sec:sd}.
Then, the chosen draft is trained over a pre-training dataset of natural language for language modeling with the objective of predicting the next token.
A common application of LLMs is code completion.
While the pre-training dataset may include some code, unless the dataset is specifically focused on code, the draft’s performance on tasks requiring an understanding of code is often suboptimal.
To address this, \cite{aryabumi2024code} proposed continued pre-training (CP), where training begins with a pre-trained model and is extended using a combination of code and natural language data.
FastDraft adopts this approach for producing drafts for code completion tasks.

\subsection{Target-draft alignment}
\label{sec:alignment}
One of the primary objectives when designing a draft model is to maximize the acceptance rate with the target model, as a higher acceptance rate directly leads to greater speedup in SD.
To closely mimic the target model's behavior in real-world scenarios, we investigate two KD strategies that expose the draft model to data samples that closely reflect those generated by the target model.

\paragraph{Strategy 1: sequence-level knowledge distillation}
\label{sec:seq-level-kd}
\cite{kim2016sequence} proposed performing KD by training the student model on sequences generated by the teacher model. 
This approach is widely used in LLM training to improve model quality \citep{alpaca, abdin2024phi}. 
To align the pre-trained draft model with the teacher model, we employ sequence-level KD by fine-tuning the draft model on a synthetic dataset generated by the target.
In contrast to the sequence-level KD approach proposed by \cite{kim2016sequence}, we differentiate between training on teacher-generated data and training with KD using the teacher’s logits. 
Accordingly, in this work, sequence-level KD refers specifically to training on generated tokens with cross-entropy loss, without incorporating the teacher’s soft targets (logits).

\paragraph{Strategy 2: token-level knowledge distillation}
\label{sec:ft-with-kd}
This strategy aligns with the traditional KD method presented in \citep{hinton2015distilling}, which involves calculating the divergence of the token level probability distribution over the vocabulary between the teacher and the draft.
Since computing the KD loss depends on the teacher's logits, the teacher is required to perform inference at every training step of the draft, making this process both computationally intensive and time-consuming.
An alternative approach is to precompute the teacher's logits beforehand and embed them into the dataset.
However, the memory demands of such a dataset can quickly exceed several terabytes, as the number of logits per token equals the vocabulary size\footnote{The vocabulary size usually exceeds 30,000 tokens} posing a substantial memory overhead.
To mitigate this issue, only a small subset of the most significant logits per token is extracted, significantly reducing the memory requirements.
This optimization results in a 6x-9x reduction in fine-tuning time without noticeable degradation in model quality.

\subsection{Draft evaluation}
\label{subsec:evaluation}
To scale both our experiments and evaluations, we created the FastDraft evaluation framework, specifically designed for assessing draft models.
The metrics and benchmarks we implemented in FastDraft evaluation are detailed in Sections~\ref{sec:metrics} and \ref{sec:benchmarks}.
Optimizing these key metrics on the proposed benchmarks is our key objective in this work.

\subsubsection{Metrics}
\label{sec:metrics}
\paragraph{Acceptance rate}
A key metric which reflects the rate at which the target model accepts the draft model's speculated tokens.
In this work, we calculate the acceptance rate (AR) \(\alpha^\gamma\) by determining the expected number of tokens accepted per block normalized by the block's size.
The formula is expressed as:

\begin{equation}
    \alpha^\gamma=\frac{1}{N}\sum_{n=0}^{N}{\frac{\mathrm{\#(accepted 
\ tokens)}}{\gamma}}
\end{equation}

where \(N\) is the number of blocks speculated by the draft model during evaluation, and \(\gamma\) is the block size.

\paragraph{Block efficiency}
The common use of SD with fixed-size blocks motivates the introduction of a more relevant efficiency metric: block efficiency, \(\tau^\gamma\).
This is defined as the expected number of accepted tokens per block.
For a given block size \(\gamma\), block efficiency \(\tau^\gamma\) serves as a more accurate measure of performance, reflecting the average rate at which tokens are accepted within each block.
\begin{equation}
    \tau^\gamma = 1 + \alpha^\gamma\times\gamma
\end{equation}

\paragraph{Wall-clock time and Memory-Bound Speedup}
\label{def:speedup}
Given the block efficiency \(\tau^\gamma\), the expected speedup achieved by applying Speculative Decoding (SD) is expressed as \(\frac{\tau^\gamma}{c\gamma + 1}\), where \(c\) represents the ratio of latencies between the draft and target models.
Since this metric is HW-dependent, it is preferable to use a HW-agnostic measure. Considering that the expected speedup occurs in a memory-bound regime, we utilize the Memory-Bound Speedup (MBSU). 
We define \(\hat{c}\) as the ratio of parameter counts between the draft and target models.
\begin{equation}
    \mathrm{MBSU} = \frac{\tau^\gamma}{\hat{c}\gamma + 1}
\end{equation}

\subsubsection{Benchmarks}
\label{sec:benchmarks}
The most commonly used benchmarks for assessing the quality of draft models, typically open-ended generation and summarization tasks, include the XSum~\citep{narayan2018xsum}, CNN-DailyMail (CNN-DM)~\citep{hermann2015cnndailymail}, and HumanEval~\citep{chen2021humaneval} datasets.
In addition to CNN-DM and HumanEval, we include in our evaluation the TinyStories (TS)~\citep{eldan2023tinystories} and Dolly~\citep{DatabricksBlog2023DollyV2} datasets, providing a more comprehensive assessment of model performance across diverse tasks.
When utilizing the TS dataset, we generate text by starting from a random position within each sample, using the preceding tokens as the input context.

\section{Experiments}
\label{sec:experiments}
In this section we report descriptions, setup and results of our experiments.
In the following experiments, we report acceptance rate (Section~\ref{sec:metrics}) results using two sampling methods, greedy and multinomial sampling decoding with temperature $T=0.6$ with block sizes $\gamma=3$ and $\gamma=5$ unless stated otherwise.
The hyper-parameters used in our draft pre-training experiments are detailed Appendix~\ref{sec:hyperparameters-datasize-exp-appendix} unless stated otherwise.

\subsection{Experimental setup}
\label{sec:setup}
We demonstrate FastDraft by training a draft model for Phi-3-mini-4k-Instruct ~\cite{abdin2024phi} target model.
Phi-3-mini, a 3.8 billion parameters LLM, was selected as our case study due to its outstanding performance across multiple open-source LLM benchmarks, while also being capable of running locally on edge devices, such as personal computers and smartphones.

\paragraph{Draft architecture}
\label{sec:draft-arch}
Exploring different architectures for draft models lies beyond the scope of this paper. Therefore, our experiments focus on drafts with the Phi-3 architecture, modified to smaller dimensions.
Specifically, we reduced the dimensions of Phi-3 to create drafts in the size of 50M and 120M parameters which are approximately 76x and 32x smaller than Phi-3-mini.
Full details of drafts configurations can be found in Table~\ref{tab:draft-arch-config} in the Appendix.

\paragraph{Pre-training datasets}
\label{sec:pre-training-datasets}
For our natural language dataset, we use a 10 billion tokens (BT) sample from the FineWeb dataset \citep{penedo2024fineweb}.
FineWeb is a widely used open-source, de-duplicated, and quality-filtered dataset, comprising 15 trillion tokens derived from 96 Common Crawl snapshots \citep{commoncrawl}. 
It has been demonstrated to yield better-performing LLMs compared to other open-source pre-training datasets such as C4, RedPajama, and The Pile (Section 3.7 in \cite{penedo2024fineweb}).
For our code dataset, we use a 10BT sample from The Stack v2 smol dataset \citep{lozhkov2024starcoder}. This variant of The Stack v2 consists of 17 commonly used programming languages, as well as a substantial collection of documentation languages, configuration languages, and configuration files. 
An overview of the 10BT dataset composition is provided in Appendix \ref{sec:code-dataset-appendix}.

\paragraph{Synthetic alignment dataset}
\label{sec:synthetic-align-data}
To perform sequence-level KD as discussed in Section~\ref{sec:seq-level-kd}, we produce an instruction fine-tuning collection using the seed instructions from several open instruction datasets: Alpaca~\citep{alpaca}, OIG-small-chip2~\citep{Nguyen2023oig} and Evol-Instruct~\citep{luo2023wizardcoder}.
We collect response sequences generated by Phi-3-mini
with greedy sampling along with multinomial sampling with temperature in \(\{0.6, 0.8, 1.0\}\) to improve the diversity of the generated sequences.
Additionally, we adopt the approach proposed by \cite{xu2024magpie}, directly soliciting instructions from the target model.
These instructions are then used to generate the corresponding responses in the same manner as previously described.

\subsection{Pre-training dataset size}
\label{sec:pre-training-dataset-size-exp}
We study the impact of the pre-training dataset size on the performance of the draft model, aiming to optimize runtime and resource usage and prevent diminishing returns.
Consequently, we uniformly sample subsets of sizes $\{0.1, 0.5, 1, 2, 5, 10\}$BT and pre-train the 50M and 120M draft configurations on these subsets.
We report the AR results for the $\{2,5,10\}$BT subsets with block size $\gamma=3$ and multinomial sampling in addition to perplexity results measured on Wikitext2~\citep{_wikitext} in Table~\ref{tab:datasize-main}.
The full results are reported in Table~\ref{tab:datasize-table} in the Appendix.

\begin{table}[t]
\centering
\resizebox{\linewidth}{!}{%
\begin{tabular}{@{}llcccc@{}}
\hline
Draft & Data size & \multicolumn{1}{l}{PPL} & CNN-DM & TS & Dolly \\ \hline
\multirow{3}{*}{50M} & 2BT & 297.4 & \textbf{0.323} & 0.264 & 0.241 \\
 & 5BT & 256.6 & 0.311 & 0.277 & \textbf{0.245} \\
 & 10BT & \textbf{240.9} & 0.312 & \textbf{0.283} & 0.234 \\ \hline
\multirow{3}{*}{120M} & 2BT & 199.6 & 0.362 & 0.297 & \textbf{0.284} \\
 & 5BT & 167.7 & \textbf{0.366} & 0.327 & 0.281 \\
 & 10BT & \textbf{147.4} & 0.351 & \textbf{0.331} & 0.251 \\ \hline
\end{tabular}
}
\caption{Pre-training data size effect on AR and perplexity (PPL). Results for block size $\gamma=3$ and multinomial sampling with temperature \(T=0.6\)}
\label{tab:datasize-main}
\end{table}

\begin{figure}[t]
    \centering
    \includegraphics[width=\linewidth, trim=0 20 0 0, clip]{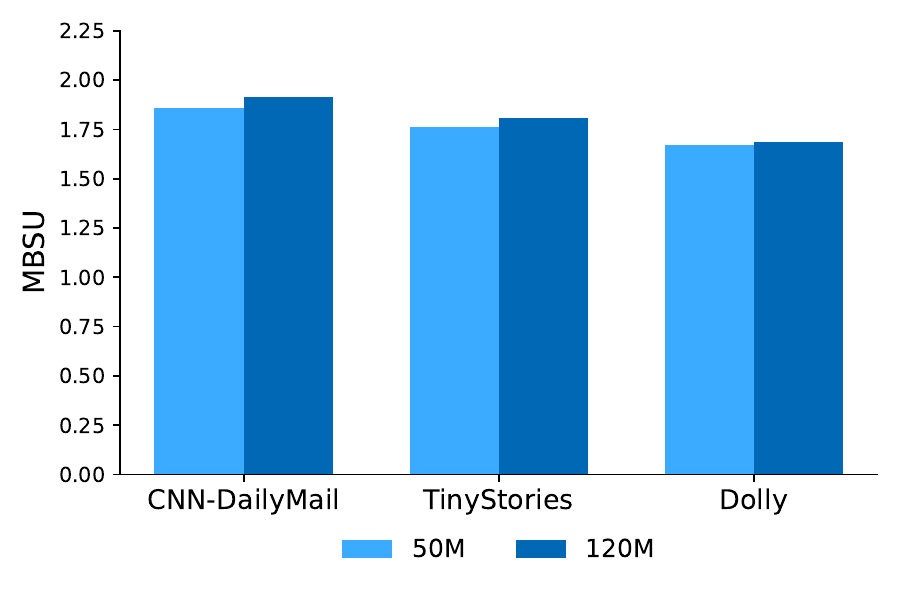}
    \caption{MBSU of 50M and 120M drafts pre-trained on 5BT FineWeb sample.}
    \label{fig:mbsu-50-vs-120}
\end{figure}

As anticipated, perplexity values decrease as the models are exposed to more training data, though the rate of improvement slows down.
However, when looking at the AR results, it is not the case.
In CNN-DM and Dolly AR either plateaus or decreases as the data size grows.
In the case of TS, AR increases with the amount of data which can be expected due to the nature of the benchmark of text completion versus instruction following and summarization.
Overall, we obtain strong results across all dataset sizes, with the models trained on the 5BT dataset emerging as a promising middle-ground option.
When comparing the 50M draft to the 120M draft, the 120M draft demonstrates superior performance in terms of AR.
Additionally, while the 120M draft also outperforms the 50M model in MBSU, the margin of improvement is relatively small, see Figure~\ref{fig:mbsu-50-vs-120}.
However, it is important to note that training the 120M draft requires twice the time compared to the 50M draft.

For the subsequent experiments, we selected the 50M draft model trained on 5BT tokens as our pre-trained draft, unless otherwise specified.

\begin{figure*}[t]
    \centering
    \includegraphics[width=\textwidth, trim=0 40 0 0, clip]{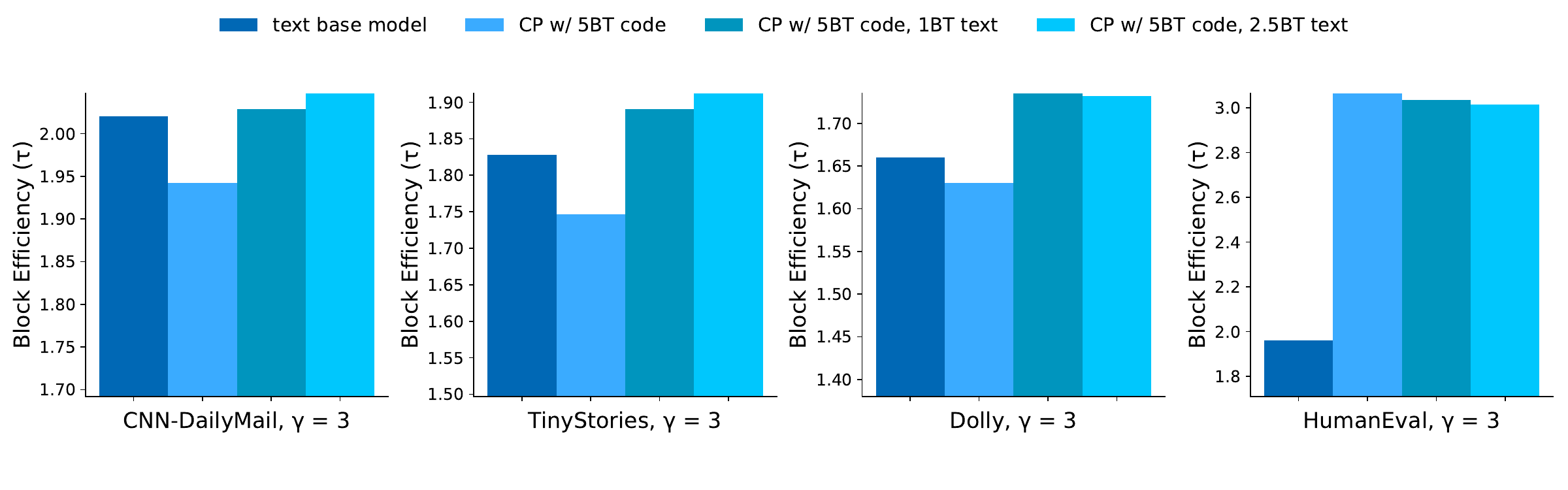}
    \caption{Block efficiency results for continued pre-training with code on tasks: CNN-DM, TS, Dolly and HumanEval using greedy decoding with block size $\gamma=3$}
    \label{fig:block-eff-win3}
\end{figure*}

\subsection{Continued pre-training for code}
\label{sec:code-pre-training-exp}
We investigate the continued pre-training (CP) method introduced by \citep{aryabumi2024code} to refine our draft model on code-related tasks. In this approach, we extend the training of our pre-trained draft model using three distinct combinations of code and natural language datasets. Each combination incorporates 5BT of code data, along with varying amounts of natural language data: specifically, ${0, 1, 2.5}$BT.
We then evaluate the resulting models on our benchmarks, including the HumanEval benchmark, and present the block efficiency results, $\tau^\gamma$, in Figure~\ref{fig:block-eff-win3}.
Our findings indicate that mixed CP datasets lead to significantly higher block efficiency in natural language tasks compared to using CP with code alone.
Additionally, block efficiency improves across natural language benchmarks, surpassing the performance of the base draft model that was trained solely on natural language, with only a slight decline observed in the Dolly benchmark.
Notably, the base draft shows a substantial improvement on the HumanEval benchmark.
The enhancement in the draft’s performance on natural language tasks with CP was anticipated, based on the findings of \citep{aryabumi2024code}. 
However, unlike the conclusions presented in that study, which advocate for a balanced dataset of natural language and code, our results suggest that CP is the preferred approach for integrating code and natural language datasets when training a draft model.

Given the structured nature of code, it serves as a highly suitable domain for generation with SD.
To investigate how to optimize pre-training for code drafts, we conducted a comprehensive ablation study.
This study involved pre-training from scratch on a mixture of code and pre-training with CP approach, where the base draft model was pre-trained on code, and CP was applied to adapt the model for natural language instead of code.
Detailed results and analysis of this ablation study are provided in Appendix~\ref{app:code-ablation}.
Overall, our findings indicate that initializing continued pre-training from a text-based model using a mixed CP dataset offers the most effective pre-training strategy for draft models, compared to other configurations tested.

\subsection{Fine tuning with synthetic data}
\label{sec:ft-syn-vs-orig}
Since our objective is to achieve the best draft alignment with the target model rather than the quality of generation, we investigate the impact of using the target's responses instead of the original dataset responses, as described in Section~\ref{sec:seq-level-kd}. 
We measure the draft's performance across several tasks. 
In this experiment, we utilize the OIG-small-chip dataset for fine-tuning.
We generate the target's responses with multinomial sampling at temperature of 0.6.
Token-level KD was not applied in this experiment.
The results are presented in Table~\ref{tab:original-vs-synthetic}. 
An analysis of the data reveals that using the target's answers leads to better draft alignment compared to using the original dataset answers across all tasks.

\begin{table}[t]
\centering
\resizebox{\linewidth}{!}{%
\begin{tabular}{llccc}
\hline
Sampling & Data source & CNN-DM & TS & Dolly \\ \hline
\multirow{3}{*}{Greedy} & None & 0.350 & 0.295 & 0.259 \\
 & Original & 0.357 & 0.282 & 0.328 \\
 & Target & \textbf{0.378} & \textbf{0.296} & \textbf{0.370} \\ \hline
\multirow{3}{*}{Multinomial} & None & 0.311 & 0.227 & 0.245 \\
 & Original & 0.328 & 0.268 & 0.311 \\
 & Target & \textbf{0.339} & \textbf{0.286} & \textbf{0.352} \\ \hline
\end{tabular}%
}
\caption{AR results for fine-tuning with original data vs synthetic data. Results for block size $\gamma=3$}
\label{tab:original-vs-synthetic}
\end{table}

\subsection{Fine tuning with knowledge distillation}
\label{sec:ft-w-kd-exp}
To further enhance the alignment of our base draft model, we incorporate KD alongside training on target-generated data, as discussed in Section~\ref{sec:ft-with-kd}. 
We experimented with various combinations of Cross Entropy (CE) loss, denoted as $\mathcal{L}_{CE}$, applied to the ground truth labels, and KD losses, using the sparse logits collected from the target model: $\mathcal{L}_{KL}$ and $\mathcal{L}_{TVD}$. 
We utilize the same dataset we generated in Section~\ref{sec:ft-syn-vs-orig} for this experiment.
The results of these experiments are presented in Table~\ref{tab:kd-kl-vs-tvd}.
Our findings indicate that KD in this setup doesn't provide significant advantage over CE loss on the target-generated data.
Although Total Variation Distance (TVD) was shown to negatively correlate with acceptance rates, in our results, it offers only slight benefit on some benchmarks over CE and Kullback–Leibler Divergence (KLD).

\begin{table}[t]
\centering
\resizebox{\linewidth}{!}{%
\begin{tabular}{llccc}
\hline
Sampling & Loss & CNN-DM & TS & Dolly \\ \hline
\multirow{5}{*}{Greedy} & $\mathcal{L}_{CE}$ & 0.378 & 0.296 & 0.370 \\
 & $\frac{1}{2}\mathcal{L}_{CE} + \frac{1}{2}\mathcal{L}_{KL}$ & 0.376 & 0.295 & 0.368 \\
 & $\frac{1}{2}\mathcal{L}_{CE} + \frac{1}{2}\mathcal{L}_{TVD}$ & 0.377 & 0.297 & 0.370 \\
 & $\mathcal{L}_{KL}$ & 0.374 & 0.294 & \textbf{0.371} \\
 & $\mathcal{L}_{TVD}$ & \textbf{0.384} & \textbf{0.301} & 0.362 \\ \hline
\multirow{5}{*}{Multinomial} & $\mathcal{L}_{CE}$ & 0.339 & 0.286 & \textbf{0.352} \\
 & $\frac{1}{2}\mathcal{L}_{CE} + \frac{1}{2}\mathcal{L}_{KL}$ & 0.350 & 0.289 & 0.346 \\
 & $\frac{1}{2}\mathcal{L}_{CE} + \frac{1}{2}\mathcal{L}_{TVD}$ & \textbf{0.356} & \textbf{0.297} & 0.347 \\
 & $\mathcal{L}_{KL}$ & 0.355 & 0.280 & 0.343 \\
 & $\mathcal{L}_{TVD}$ & 0.347 & \textbf{0.297} & 0.344 \\ \hline
\end{tabular}%
}
\caption{KD with KL-Divergence and TVD. Results for block size $\gamma=3$}
\label{tab:kd-kl-vs-tvd}
\end{table}

\begin{table*}[t]
\centering
\resizebox{\textwidth}{!}{%
\begin{tabular}{@{}llcccccccc@{}}
\hline
\multirow{2}{*}{Model} & \multirow{2}{*}{Type} & \multicolumn{2}{c}{CNN-DM} & \multicolumn{2}{c}{TS} & \multicolumn{2}{c}{Dolly} & \multicolumn{2}{c}{HumanEval}\\
 &  & $\gamma=3$ & $\gamma=5$ & $\gamma=3$ & $\gamma=5$ & $\gamma=3$ & $\gamma=5$ & $\gamma=3$ & $\gamma=5$ \\ \hline
\multirow{3}{*}{Phi3-mini 50M} & \( PT\) & 0.311 & 0.221 & 0.277 & 0.184 & 0.245 & 0.163 & 0.229 & 0.151\\
 & \( PT \rightarrow CP \) & 0.304 & 0.211 & 0.287 & 0.192 & 0.226 & 0.149 & 0.561 & 0.450\\
 & \( PT \rightarrow CP \rightarrow FT\) & \textbf{0.369} & \textbf{0.267} & \textbf{0.306} & \textbf{0.208} & \textbf{0.370} & \textbf{0.265} & \textbf{0.562} & \textbf{0.472}\\ \hline
\multirow{3}{*}{Llama3.1 150M} & \( PT\) & 0.280 & 0.186 & 0.227 & 0.147 & 0.247 & 0.158 & 0.248 & 0.168\\
 & \( PT \rightarrow CP \) & 0.280 & 0.192 & 0.235 & 0.155 & 0.273 & 0.176 & 0.606 & 0.480\\
 & \( PT \rightarrow CP \rightarrow FT\) & \textbf{0.307} & \textbf{0.214} & \textbf{0.266} & \textbf{0.178} & \textbf{0.334} & \textbf{0.239} & \textbf{0.649} & \textbf{0.525}\\
 \hline
\end{tabular}%
}
\caption{AR results for FastDraft stages, performed in subsequent order: Pre-training (PT), Continued Pre-training (CP) and Fine-tuning (FT) using multinomial sampling with temperature \(T=0.6\)}
\label{tab:full-pipeline-multinomial}
\end{table*}

\subsection{Hardware-aware draft}
\label{subsec:ha-selection}
The high quality of models like Phi-3-mini and Llama-3.1-8B, combined with advances in edge device hardware, has made local deployment of LLMs increasingly appealing, yet still challenging. 
Typical use cases for locally deployed LLMs share similar characteristics, they are designed for single-user interactions where only one query is processed at a time.
This setup presents an excellent opportunity to apply speculative decoding to accelerate LLM generation with minimal resource demands.

The Phi3-mini 50M uses the original Transformers dimensions, however, due to the strict latency requirements of draft models, their architectural design warrants special attention, as their performance may vary depending on the hardware platform.
We conduct a series of experiments to examine the performance of draft models derived from different architectural choices.
Our focus is on altering the depth of the models by changing the number of layers or the width of the models by modifying the hidden dimension.
The target system for our experiments is the latest
{Intel\textsuperscript{\tiny\textregistered} Core\textsuperscript{\tiny\texttrademark} Ultra 7 Processor 268V}
coupled with Intel\textsuperscript{\tiny\textregistered} OpenVINO\textsuperscript{\tiny\texttrademark} runtime engine.

\begin{figure}[t]
    \centering
    \includegraphics[width=\linewidth]{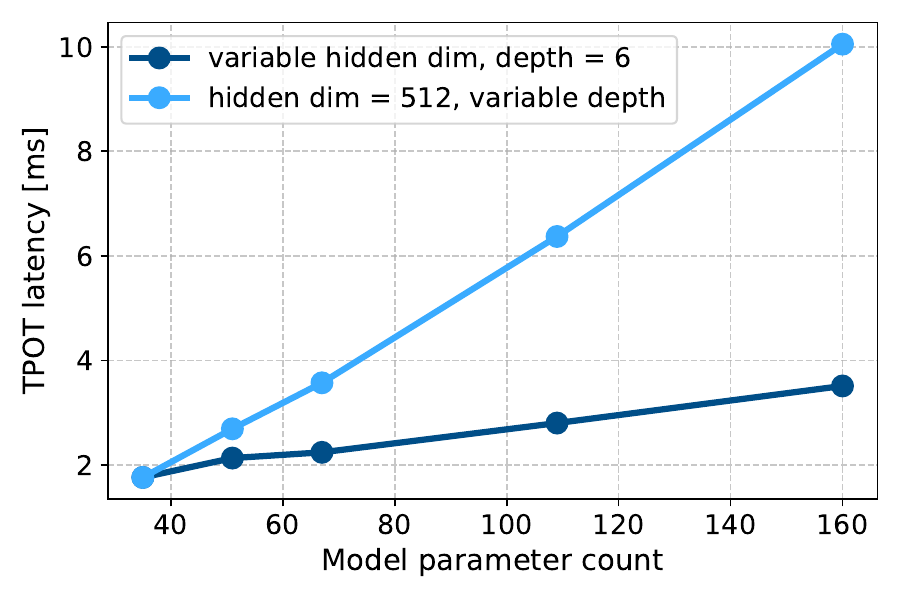}
    \caption{Latency of models obtained by varying either a hidden dimension or a number of layers}
    \label{fig:latency-layers-vs-hidden}
\end{figure}

First, we analyze the effect of scaling the width and depth of the draft on latency of the draft model on the target hardware.
We fix either the hidden dimension or the model depth to match the corresponding values of Phi3-mini 50M draft and modify the other parameter to observe how latency scales with wider or deeper models.
As shown in  Figure~\ref{fig:latency-layers-vs-hidden}, our benchmarking results suggest that increasing the number of layers has the most significant impact on latency, while increasing the hidden dimension leads to only a slight increase.

\begin{figure}[t]
    \centering
    \includegraphics[width=\linewidth]{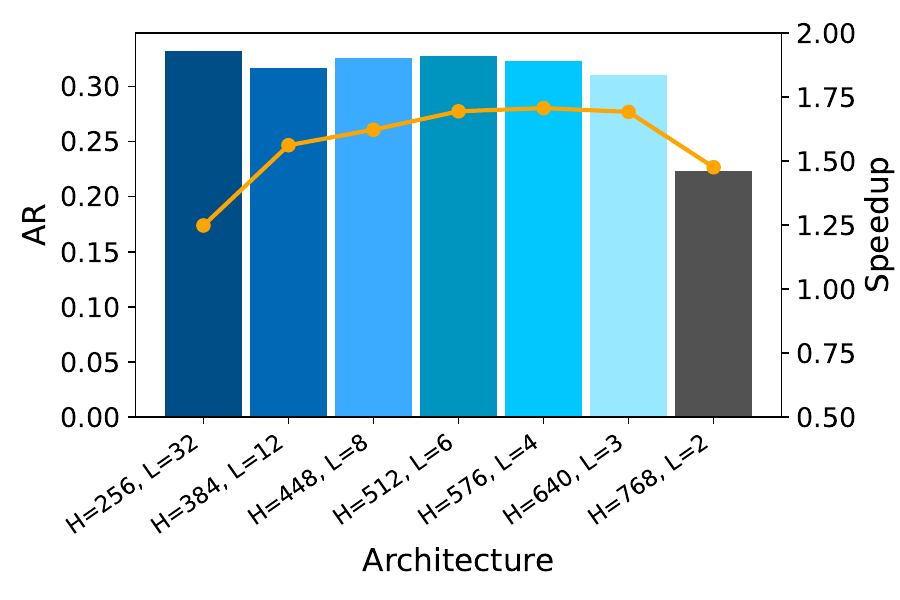}
    \caption{Estimated speedup on CNN-DM for models with constant parameter budget. L, H denote the number of layers and hidden dimension size}
    \label{fig:constant_budget}
\end{figure}

Next, we evaluate the impact of the draft architecture under a fixed parameter budget\footnote{We excluded embedding table size from the model parameter budget, as it does not strain memory bandwidth.} by assessing the AR and estimated speedup achieved when varying the width and depth of the draft.
By redistributing parameters between these dimensions, we aimed to identify configurations that optimize the trade-off between speculation quality and computational efficiency.
We pre-trained multiple draft models for Phi-3-mini, varying the number of layers and hidden dimension sizes while maintaining a fixed parameter count.
We then measured their AR on a target dataset and recorded their latency on our target hardware to estimate the speedup achieved through SD via Equation~\ref{eq:sd-speedup}.
Figure~\ref{fig:constant_budget} presents the results for the CNN-DM dataset.
Our findings indicate that increasing depth does not improve the AR, while excessively shallow drafts degrade the performance.
Moreover, the AR–latency trade-off follows an inverted U-shaped speedup curve with a broad base, consistent with our earlier experimental results.
The results for the TS and Dolly datasets, presented in Appendix~\ref{sec:hardware-aware-details-appendix}, exhibited the same pattern.

\section{Results \& reproducibility}
\label{sec:final-recipe}
Combining the findings from the ablation studies presented in Section~\ref{sec:experiments}, we outline our comprehensive pipeline for draft pre-training and fine-tuning to optimize performance on key metrics for speculative decoding and demonstrate it by producing a draft for Phi-3-mini. 
To further illustrate reproducibility, we employ the same pipeline to produce a draft for Llama-3.1-8B-Instruct \cite{dubey2024llama}.
We present a 50M draft for the Phi-3-mini and a 150M draft for Llama-3.1-8B. Drafts' architecture details are presented in Appendix~\ref{app:arch-details}. 
Considering both performance and resource efficiency, our findings from sections \ref{sec:pre-training-dataset-size-exp}, \ref{sec:code-pre-training-exp} suggest pre-training over 5BT of FineWeb text data and continued pre-training on a mixture of 5BT code data from The Stack v2  and  2.5BT FineWeb text data as a best practice.
For fine-tuning, we conclude from sections \ref{sec:ft-syn-vs-orig}, \ref{sec:ft-w-kd-exp} that sequence-level KD yields significant improvements while token-level KD benefits are not definitive, therefore, for FastDraft we only utilize sequence-level KD.
We construct an alignment dataset for both Phi-3 and Llama-3.1 drafts by combining a number of synthetic datasets we generated with the appropriate target model as described in Section~\ref{sec:synthetic-align-data}.
Table~\ref{tab:full-pipeline-multinomial} presents the AR improvements achieved through each stage of FastDraft generated with multinomial sampling.
Results for greedy sampling are presented in Table~\ref{tab:full-pipeline-greedy} in the Appendix.
Using both sampling methods, we observe for datasets CNN-DM, TS, and Dolly, a substantial AR increase following fine tuning, with instruction-following dataset Dolly exhibiting an increase of  \textasciitilde10\% in AR. 
For HumanEval the primary performance gains stem from pre-training on code domain knowledge during continued pre-training.
These models achieve a MBSU of \textasciitilde2x for natural language tasks and \textasciitilde3x for code completion tasks.

Furthermore, we demonstrate the real-world wall-clock time speedup achieved with FastDraft on the {Intel\textsuperscript{\tiny\textregistered} Core\textsuperscript{\tiny\texttrademark} Ultra 7 Processor 268V}
as detailed in Section~\ref{subsec:ha-selection}.
We observe an average TPOT speedup of 1.5x for summarization tasks and 2x for code completion. 
Additionally, memory transaction efficiency improves, with up to 3x reduction in memory bandwidth usage, as indicated by the MBSU metric in these trials.
Memory transactions can be highly energy-intensive \citep{horowitz20141}, making the use of FastDraft drafts a promising approach for achieving substantial power savings on edge devices. 
These performance gains stem from the compact nature of our drafts, enabling them to operate in tandem with large target LLMs.
Notably, a pre-trained FastDraft draft model can be used either as-is or aligned with any compatible model.
For instance, our Llama-3.1 draft is readily compatible with over 1,000 models on the HuggingFace Hub\footnote{\url{https://huggingface.co/}}, highlighting FastDraft's scalability.
We believe FastDraft has the potential to unlock LLMs inference on AI-PC and other edge-devices.

\section{Conclusion \& future work}
In this paper, we introduced FastDraft, a novel approach for training and evaluating draft models for speculative decoding.
Our results demonstrate that FastDraft facilitates the rapid training of high-quality, efficient draft models that are well-aligned with target models.
We conducted a comprehensive ablation study to examine various aspects of draft training, including pre-training data composition and draft-target alignment.
Using our method, we successfully trained a highly efficient 50M-parameter draft model for Phi-3-mini, achieving an acceptance rate of up to 67\% and up to a 3x memory-bound speedup.
We further demonstrated the scalability of FastDraft by training a draft for Llama-3.1-8B-Instruct, underscoring  its effectiveness for larger models.
We validated our theoretical findings by benchmarking our drafts on the latest AI-PC HW achieving a wall-clock time speedup of up to 2x.
We hope our findings will inspire additional research on efficient draft training, particularly focusing on the development of resource-efficient draft architectures and hardware-aware designs.

\section{Limitations}
Our training recipe for draft models has been validated exclusively in English, and the reported acceptance rates and speedup benefits may not generalize to non-English languages due to their distinct syntactic structures and linguistic characteristics.
Future work should investigate the effectiveness of our approach across different languages and develop language-specific modifications to optimize performance for non-English applications.

While prior work has demonstrated that speculative decoding with multiple candidate sequences can achieve higher acceptance rates and greater speedup compared to single-sequence approaches \citep{li2024eagle, cai2024medusa}, we restricted our implementation to single-sequence speculation due to computational constraints.
Multiple-sequence speculation requires substantial parallel compute resources to generate and validate multiple candidate sequences simultaneously.
Given our focus on AI-PC, we prioritized developing methods that could run effectively on consumer HW.

Our work primarily investigates optimal training strategies for draft models that share the same architectural design as their target models.
While this focused scope allowed us to deeply explore the relationship between architecturally identical draft and target pairs, it also presents notable limitations.
First, our findings may not generalize well to scenarios where draft and target models employ different architectures, as the dynamics of knowledge transfer and performance optimization could vary significantly in such cases.
Additionally, while our proposed training recipe proved effective for architecturally identical pairs, it's possible that combining our approach with more efficient draft architectures could yield even better results—potentially offering improved speed-quality trade-offs that we have not explored in this work.

\bibliography{custom}

\clearpage
\appendix
\section{Comparison to other methods}
\label{app:eagle-comparison}
To ensure a fair comparison with speculative decoding methods such as MEDUSA~\citep{cai2024medusa}, EAGLE~\citep{li2024eagle}, and self-speculation techniques as presented in~\citep{zhang2023draft_n_verify,xia2024swift}, key factors must remain consistent, including hardware, target model, and speculation strategy. 
In contrast to the aforementioned methods, FastDraft adheres strictly to standard, sequential speculative decoding without employing multi-token windows, multiple drafts, or speculation trees—although such enhancements are complementary to our approach and could be incorporated to achieve additional speed improvements. 
Moreover, our primary objective is to offer a solution in scenarios where utilizing an existing draft model or a smaller language model (SLM) from the same model family is not feasible (e.g., Phi-3). 

As reported in \citealp{goel2024direct}, pretrained draft models have been empirically shown to exhibit significantly better alignment with the target model. 
Despite the associated cost, our approach enables efficient training from scratch in several key aspects:
FastDraft supports architectural flexibility, enabling the construction of compact draft models. 
In contrast, drafts designed to leverage latent representations of the target model scale with the target's hidden dimension. 
For example, EAGLE drafts for 7B, 13B, and 70B models require 0.24B, 0.37B, and 0.99B parameters, respectively, while MEDUSA's 3–5 decoding heads correspond to 1.5–2.6B trainable parameters for Llama3.1-8B. 
In the case of Self-Speculation, the large budget of target parameters—approximately 500M for Llama-3.1-8B—used during speculation also limits speedup and results in substantial memory consumption.
The FastDraft pipeline can be completed within one day using eight accelarators, representing a substantial improvement over prior draft training methods ~\citep{goel2024direct,miao2023specinfer}.
For comparison, EAGLE training requires up to 2 days on four GPUs.
Furthermore, EAGLE-3 \citep{li2025eagle3} was trained on approximately eight times more data than EAGLE. 
Moreover, FastDraft drafts can be used with any compatible target model without additional training or adjustments, whereas fine-tuning can further improve alignment to the target model.
Self-Speculation, by contrast, depends on Bayesian optimization—a computationally intensive process that can take several hours per iteration.

\section{Pre-training details and results}
\label{sec:pre-training-details-appendix}
\subsection{Draft model architecture}
\label{app:arch-details}
We work with drafts built on the architecture of "Phi-3-mini-4k-instruct" and "Llama-3.1-8B-Instruct".
We use float16 precision for model pre-training. Table \ref{tab:draft-arch-config} below provides a detailed view of the structures of the drafts.

\begin{table*}[h]
\centering
\begin{tabular}{@{}lccc@{}}
\hline
Draft name & Phi3-mini 50M & Phi3-mini 120M & Llama3.1 150M\\ \hline
Hidden size & 512 & 768 & 512 \\
Intermediate size & 1408 & 2048 & 1792 \\
\# Layers & 6 & 12 & 6 \\
\#Attention heads & 8 & 12 & 8 \\
\# Key-value heads & 8 & 12 & 8 \\
Vocabulary size & 32064 & 32064 & 128256 \\ \hline
\end{tabular}
\caption{Phi-3-mini and Llama-3.1-8B-Instruct drafts configurations}
\label{tab:draft-arch-config}
\end{table*}

\subsection{Details of pre-training code dataset}
\label{sec:code-dataset-appendix}
In Table \ref{codedata-table}, we summarize the data composition of the code dataset we employ for pre-training.

\begin{table*}[t]
    \renewcommand{\arraystretch}{1.2}
    \begin{tabular}{llll}
        \hline
        Language     & Token count (B) & Sample Count & Avg. sample length (tokens) \\
        \hline
        Python & 1.18 & 963923 & 1227 \\
        C & 0.83 & 309496 & 2679 \\
        C\# & 0.78 & 783399 & 994 \\
        C++ & 1.35 & 676943 & 1994 \\
        Go & 0.18 & 140831 & 1285 \\
        Java & 1.02 & 1003702 & 1015 \\
        JavaScript & 1.27 & 1109269 & 1145 \\
        Kotlin & 0.09 & 142275 & 601 \\
        Lua & 0.1 & 32335 & 3212 \\
        PHP & 1.07 & 731103 & 1462 \\
        R & 0.13 & 79108 & 1704 \\
        Ruby & 0.13 & 285785 & 469 \\
        Rust & 0.07 & 36811 & 1926 \\
        SQL & 0.26 & 70937 & 3633 \\
        Shell & 0.11 & 175985 & 613 \\
        Swift & 0.1 & 118558 & 877 \\
        TypeScript & 0.26 & 386644 & 674 \\
        Documentation languages & 0.96 & 807360 & 1194 \\
        Configuration languages & 0.09 & 140370 & 652 \\
        Configuration files & 0.02 & 43507 & 554 \\
        \hline
        Total & 10.0 &  8038341 &  \\
        \hline
    \end{tabular}
    \caption{Overview of the data composition of the 10BT sample of the-stack-v2-train-smol}
    \label{codedata-table}
\end{table*}

\subsection{Hyper-parameter configuration for datasets size in pre-training experiments}
\label{sec:hyperparameters-datasize-exp-appendix}
We train all draft variants, for both pre-training and continued pre-training (CP) in float16 precision for 1 epoch with a batch size of 128. 
We use Adam with learning rate of $1 \times 10^{-4}$, $\beta_1 = 0.9$, $\beta_2 = 0.999$, L2 weight decay of 0.01, learning rate warmup over 5\% of the total training steps, 
and linear decay of the learning rate.

\subsection{Pre-training model size ablation}
\label{app:data-size}

We provide the full results of Table \ref{tab:datasize-main}, pre-training on varying model sizes of 50 and 120 million parameters in Table \ref{tab:datasize-table}. For a detailed view of the drafts' structure, see Table \ref{tab:draft-arch-config}

\begin{table*}[t]
\centering
\resizebox{\textwidth}{!}{%
\begin{tabular}{@{}lllcccccc@{}}
\hline
\multirow{2}{*}{Draft size} & \multirow{2}{*}{Data size} & \multirow{2}{*}{Perplexity} & \multicolumn{2}{c}{CNN-DM} & \multicolumn{2}{c}{TS} & \multicolumn{2}{c}{Dolly} \\
 &  &  & Greedy & Multinomial & Greedy & Multinomial & Greedy & Multinomial \\ \hline
\multirow{6}{*}{50M} & 0.1BT & 1594.5 & 0.168 & 0.152 & 0.158 & 0.143 & 0.173 & 0.151 \\
 & 0.5BT & 476.3 & 0.304 & 0.280 & 0.245 & 0.226 & 0.253 & 0.233 \\
 & 1BT & 379.0 & 0.315 & 0.295 & 0.267 & 0.247 & 0.255 & 0.236 \\
 & 2BT & 297.4 & 0.347 & \textbf{0.323} & 0.267 & 0.247 & 0.255 & 0.236 \\
 & 5BT & 256.6 & \textbf{0.350} & 0.311 & 0.295 & 0.277 & \textbf{0.258} & \textbf{0.245} \\
 & 10BT & \textbf{240.9} & 0.325 & 0.312 & \textbf{0.300} & \textbf{0.283} & 0.242 & 0.234 \\ \hline
\multirow{6}{*}{120M} & 0.1BT & 1241.0 & 0.194 & 0.180 & 0.183 & 0.171 & 0.199 & 0.173 \\
 & 0.5BT & 343.2 & 0.322 & 0.302 & 0.279 & 0.260 & 0.286 & 0.266 \\
 & 1BT & 253.8 & 0.376 & 0.348 & 0.302 & 0.272 & \textbf{0.314} & \textbf{0.294} \\
 & 2BT & 199.6 & 0.391 & 0.362 & 0.321 & 0.297 & 0.295 & 0.284 \\
 & 5BT & 167.7 & \textbf{0.393} & \textbf{0.366} & 0.343 & 0.327 & 0.293 & 0.281 \\
 & 10BT & \textbf{147.4} & 0.381 & 0.351 & \textbf{0.357} & \textbf{0.331} & 0.272 & 0.251 \\ \hline
\end{tabular}%
}
\caption{Pre-training data size effect on acceptance rate of natural language tasks. Results for block size $\gamma=3$ on both greedy and multinomial sampling temperature $T=0.6$}
\label{tab:datasize-table}
\end{table*}


\subsection{Pre-training on code ablation}
\label{app:code-ablation}

\paragraph{Pre-training on a mixed dataset}
\label{sec:mixed-pre-training-exp}
Consistent with current best practices for pre-training large language models \cite{dubey2024llama,abdin2024phi},
we experiment with pre-training the draft model on a mixture of code and natural language datasets with varying ratios of the domains.
Similar to \cite{aryabumi2024code}, we evaluate a balanced mixture of 5 billion token code and 5 billion token text datasets.
Since our datasets are relatively small to begin with, emphasizing on text data can be beneficial for general language understanding across tasks. Accordingly, we also experiment with mixed datasets featuring 4 billion tokens of text and 6 billion tokens of code, as well as 7 billion tokens of text and 3 billion tokens of code. 

Our experimental results suggest the following conclusions:
\paragraph{Continued pre-training is better than pre-training on a mixed dataset:} Continued pre-training variants with CP mixed dataset of code and text were able to obtain 1-4\% higher acceptance rate on CNN-DM, TS and Dolly datasets, over greedy and multinomial sampling methods (see Table \ref{tab:greedy-CP-natural-lang}, \ref{tab:stochastic-CP-natural-lang}) compared to CP datasets of only code or only text. These models have been trained on 1-2.5 billion more tokens compared to pre-trained models on mixed data, but as section \ref{sec:pre-training-dataset-size-exp} suggests, the performance gain likely stems from the choice of pre-training strategy rather than additional data. Surprisingly, mixed datasets with higher proportion of text yield better results on HumanEval, a code evaluation dataset, with little to no improvement on natural language tasks \ref{tab:greedy-mix}, \ref{tab:stochastic-mix}. This strengths our findings in section \ref{sec:pre-training-dataset-size-exp} that additional text data doesn't contribute to our draft performance beyond a certain point.

\paragraph{Text Base dataset is better than code Base dataset for continued pre-training} Variants of continued pre-training settings with Base dataset of Fineweb 5 billion token sample achieve higher acceptance rate on HumanEval (Table \ref{tab:greedy-CP-code}) compared to their code Base dataset counterparts of TheStack-v2 5 billion token sample. 
This is likely because these models acquire the majority of their code domain knowledge during the initial pre-training stage, and the later introduction of text domain knowledge can cause gradient shifts that are sub-optimal for code. While including portions of code in the CP dataset helps mitigate this issue, the performance degrades significantly by up to 50\% when the CP dataset lacks code.

 \begin{table*}[t]
    \renewcommand{\arraystretch}{1.2}
    \centering
    \begin{tabular}{ccccccc}
    \hline
    \multicolumn{1}{c}{\multirow{2}{*}{Base dataset}} & \multicolumn{1}{c}{CP mix dataset} & \multicolumn{1}{c}{CP mix dataset} & \multicolumn{2}{c}{HumanEval} \\
    & code & natural lang. & $\gamma=3$ & $\gamma=5$ \\
    \hline
    \vspace{-0.5cm}
    \multirow{3}{*}{\begin{tabular}{c} Fineweb \\ \vspace{-0.1cm} 5BT sample \end{tabular}} \\
    & 5B & - & \textbf{0.688} & \textbf{0.578} \\
    & 5B & 1B & 0.678 & 0.560 \\
    & 5B & 2.5B & 0.672 & 0.561 \\
    \hline
    \vspace{-0.5cm}
    \multirow{3}{*}{\begin{tabular}{c} The stack-v2 \\ \vspace{-0.1cm} 5BT sample \end{tabular}} \\
    & - & 5B & 0.320 & 0.238 \\
    & 1B & 5B & 0.628 & 0.519 \\
    & 2.5B & 5B & 0.667 & 0.553 \\
    \hline
    \end{tabular}
    \caption{Continued pre-training effect on acceptance rate of code using greedy decoding}
    \label{tab:greedy-CP-code}
\end{table*}

\begin{table*}
    \renewcommand{\arraystretch}{1.3}
    \centering
\resizebox{\textwidth}{!}{
    \begin{tabular}{ccccccccccc}
    \hline
    \multicolumn{1}{c}{\multirow{2}{*}{CP Base dataset}} & \multicolumn{1}{c}{CP dataset} & \multicolumn{1}{c}{CP dataset} & \multicolumn{2}{c}{CNN-DM} & \multicolumn{2}{c}{TS} & \multicolumn{2}{c}{Dolly} \\
    \cline{4-9}
    & code & natural lang.  & $\gamma=3$ & $\gamma=5$ & $\gamma=3$ & $\gamma=5$ & $\gamma=3$ & $\gamma=5$  \\
    \hline
    \vspace{-0.5cm}
    \multirow{3}{*}{\begin{tabular}{c} Fineweb \\ \vspace{-0.1cm} 5BT sample \end{tabular}} \\
    & 5B & - & 0.314 & 0.216 & 0.249  & 0.164 & 0.210 & 0.139 \\
    & 5B & 1B & 0.343 & 0.241 & 0.297 & 0.200 & \textbf{0.245} & \textbf{0.163} \\
    & 5B & 2.5B & \textbf{0.349} & \textbf{0.244} & \textbf{0.304}  & \textbf{0.204} & 0.244 & \textbf{0.163} \\
    \hline
    \vspace{-0.5cm}
    \multirow{3}{*}{\begin{tabular}{c} The stack-v2 \\ \vspace{-0.1cm} 5BT sample \end{tabular}} \\
    & - & 5B & 0.340 & 0.238 & 0.276  & 0.182 & 0.220 & 0.144 \\
    & 1B & 5B & 0.339 & 0.239 & 0.283  & 0.190 & 0.222 & 0.148 \\
    & 2.5B & 5B & 0.344 & 0.243 & 0.285 & 0.191 & 0.227 & 0.151  & \\
    \hline
    \end{tabular}    
    }
    \caption{Continued pre-training effect on acceptance rate of natural language tasks using greedy decoding}
    \label{tab:greedy-CP-natural-lang}
\end{table*}

\begin{table*}
    \renewcommand{\arraystretch}{1.2}
    \centering
    \begin{tabular}{ccccccc}
    \hline
    \multicolumn{1}{c}{\multirow{2}{*}{Base dataset}} & \multicolumn{1}{c}{CP mix dataset} & \multicolumn{1}{c}{CP mix dataset} & \multicolumn{2}{c}{HumanEval} \\
    \cline{4-5}
    & code & natural lang.  & $\gamma=3$ & $\gamma=5$ \\
    \hline
    \vspace{-0.5cm}
    \multirow{3}{*}{\begin{tabular}{c} Fineweb \\ \vspace{-0.1cm} 5BT sample \end{tabular}} \\
    & 5B & - & \textbf{0.578} & \textbf{0.462} \\
    & 5B & 1B & 0.560 & 0.451 \\
    & 5B & 2.5B & 0.561 & 0.450 \\
    \hline
    \vspace{-0.5cm}
    \multirow{3}{*}{\begin{tabular}{c} The stack-v2 \\ \vspace{-0.1cm} 5BT sample \end{tabular}} \\
    & - & 5B & 0.238 & 0.161 \\
    & 1B & 5B & 0.519 & 0.400 \\
    & 2.5B & 5B & 0.553 & 0.416 \\
    \hline
    \end{tabular}
    \caption{Continued pre-training effect on acceptance rate of code using multinomial sampling with temperature $T=0.6$}
    \label{tab:stochastic-CP-code}
\end{table*}

\begin{table*}
    \renewcommand{\arraystretch}{1.3}
    \centering
    \resizebox{\textwidth}{!}{
    \begin{tabular}{ccccccccccc}
    \hline
    \multicolumn{1}{c}{\multirow{2}{*}{CP Base dataset}} & \multicolumn{1}{c}{CP mix dataset} & \multicolumn{1}{c}{CP mix dataset} & \multicolumn{2}{c}{CNN-DM} & \multicolumn{2}{c}{TS} & \multicolumn{2}{c}{Dolly} \\
    \cline{4-9}
    & code & natural lang. & $\gamma=3$ & $\gamma=5$ & $\gamma=3$ & $\gamma=5$ & $\gamma=3$ & $\gamma=5$  \\
    \hline
    \vspace{-0.5cm}
    \multirow{3}{*}{\begin{tabular}{c} Fineweb \\ \vspace{-0.1cm} 5BT sample \end{tabular}} \\
    & 5B & - & 0.271 & 0.170 & 0.232 & 0.153 & 0.191 & 0.126  \\
    & 5B & 1B & 0.296 & 0.210 & 0.279 & 0.187 & \textbf{0.233} & 0.148  \\
    & 5B & 2.5B & 0.304 & 0.211 & \textbf{0.287} & \textbf{0.192} & 0.226 & \textbf{0.149}\\
    \hline
    \vspace{-0.5cm}
    \multirow{3}{*}{\begin{tabular}{c} The stack-v2 \\ \vspace{-0.1cm} 5BT sample \end{tabular}} \\
    & - & 5B & 0.297 & 0.205 & 0.254 & 0.173 & 0.201 & 0.135 \\
    & 1B & 5B & 0.307 & 0.201 & 0.268 & 0.172 & 0.205 & 0.141 \\
    & 2.5B & 5B & \textbf{0.312} & \textbf{0.213} & 0.264 & 0.176 & 0.215 & 0.143 \\
    \hline
    \end{tabular}  
    }
    \caption{Continued pre-training effect on acceptance rate of natural language tasks using multinomial sampling with temperature $T=0.6$}
    \label{tab:stochastic-CP-natural-lang}
\end{table*}

\begin{table*}
    \renewcommand{\arraystretch}{1.2}
    \centering
\resizebox{\textwidth}{!}{
    
    \begin{tabular}{cccccccccccc}
    \hline
    \multicolumn{1}{c}{Mix dataset} & \multicolumn{1}{c}{Mix dataset} & \multicolumn{2}{c}{CNN-DM} & \multicolumn{2}{c}{TS} & \multicolumn{2}{c}{Dolly} & \multicolumn{2}{c}{HumanEval} \\
    \cline{3-10}
    natural lang. & code & $\gamma=3$ & $\gamma=5$ & $\gamma=3$ & $\gamma=5$ & $\gamma=3$ & $\gamma=5$ & $\gamma=3$ & $\gamma=5$ \\
    \hline
    5B & 5B & 0.354 & 0.249 & 0.284 & 0.189 & 0.205 & 0.136 & 0.639 & 0.532 \\
    6B & 4B & 0.355 & 0.248 & 0.284 & 0.190 & \textbf{0.220} & \textbf{0.144} & 0.656 & \textbf{0.551} \\
    7B & 3B & 0.354 & 0.248 & 0.285 & 0.191 & 0.216 & \textbf{0.144} & \textbf{0.657} & 0.550 \\
    \hline
    \end{tabular}    
    }
    \caption{Impact of pre-training on mixed datasets on acceptance rate of natural language and code tasks. Evaluated across window sizes 3 and 5 using greedy decoding}
    \label{tab:greedy-mix}
\end{table*}

\begin{table*}
    \renewcommand{\arraystretch}{1.2}
    \centering
\resizebox{\textwidth}{!}{
    
    \begin{tabular}{cccccccccccc}
    \hline
    \multicolumn{1}{c}{Mix data} & \multicolumn{1}{c}{Mix data} & \multicolumn{2}{c}{CNN-DM} & \multicolumn{2}{c}{TS} & \multicolumn{2}{c}{Dolly} & \multicolumn{2}{c}{HumanEval} \\
    \cline{3-10}
    natural lang. & code & $\gamma=3$ & $\gamma=5$ & $\gamma=3$ & $\gamma=5$ & $\gamma=3$ & $\gamma=5$ & $\gamma=3$ & $\gamma=5$ \\
    \hline
    5B & 5B & 0.311	& 0.206 &	\textbf{0.265} &	0.175 &	0.194 &	0.124 &	0.506 &	0.389\\
    6B & 4B & 0.317 &	\textbf{0.218} & 0.261 & \textbf{0.178} &	0.201 & 0.137 & 0.530 & 0.411\\
    7B & 3B & \textbf{0.326} &	0.211 & 0.258 & 0.176 &	\textbf{0.205} &	\textbf{0.137} & \textbf{0.533} & \textbf{0.424} \\
    \hline
    \end{tabular}    
    }
    \caption{Impact of pre-training on mixed datasets on acceptance rate of natural language and code tasks. Evaluated across window sizes 3 and 5 using multinomial sampling with temperature $T=0.6$}
    \label{tab:stochastic-mix}
\end{table*}

\section{Hardware-aware draft details and results}
\label{sec:hardware-aware-details-appendix}
Tables~\ref{tab:latency-draft-arch-config} and~\ref{tab:latency-vs-acceptance-draft-arch-config} contain the full list of configuration used in HW-aware experiments. Table~\ref{tab:ar-vs-speedup-const-budget} lists the configurations used for estimating the speedup of architectures with varying depth or width satisfying a constant parameter budget.

\begin{figure*}
\centering
\begin{minipage}{.5\textwidth}
  \centering
  \includegraphics[width=\linewidth]{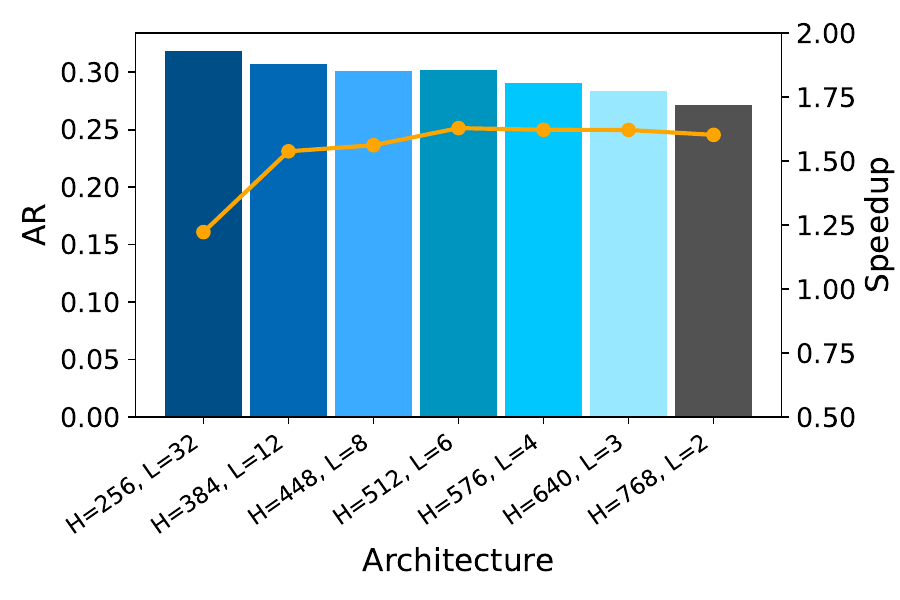}
  \captionof{figure}{Estimated speedup on TS for models with \\ constant parameter budget. L, H denote the number \\ of layers and hidden dimension size}
  \label{fig:ts}
\end{minipage}%
\begin{minipage}{.5\textwidth}
  \centering
  \includegraphics[width=\linewidth]{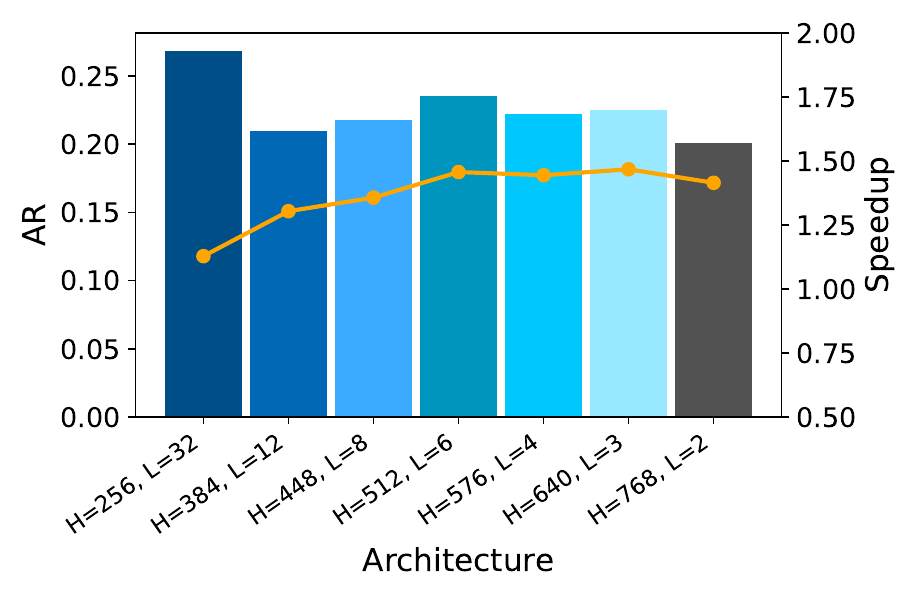}
  \captionof{figure}{Estimated speedup on Dolly for models with \\ constant parameter budget. L, H denote the number \\ of layers and hidden dimension size}
  \label{fig:dolly}
\end{minipage}
\end{figure*}

Figures~\ref{fig:ts} and~\ref{fig:dolly} present the results for the constant budget pre-trained drafts evaluated on the TS and Dolly datasets.

\begin{table*}[h]
\centering
\begin{tabular}{@{}cccccc@{}}
\hline
\# Params (M) & Hidden size & Intermediate size & \# Layers & \#Attention heads & \# Key-value heads\\ 
\hline
27.91 &	512	& 1408 & 4 & 8	& 8 \\
38.95 &	640	& 1792 & 4 & 10	& 10 \\
37.10 &	512	& 1408 & 7 & 8	& 8 \\ 
42.67 &	768	& 2048 & 4 & 12	& 12 \\
43.23 &	512	& 1408 & 9 & 8 & 8 \\
80.32 & 1024 & 2816 & 4	& 16 & 16 \\
80.00 &	512	& 1408 & 21 & 8 & 8 \\
114.78 & 1280 & 3456 & 4& 20 & 20 \\
113.70 & 512 & 1408	& 32 & 8 & 8 \\ 
\hline
34.04 &	512	& 1408 & 6 & 8 & 8 \\
48.64 &	640	& 1792 & 6 & 10 & 10 \\
49.36 &	512	& 1408 & 11 & 8 & 8 \\ 
64.00 &	768	& 2048 & 6 & 12 & 12 \\
64.68 &	512	& 1408 & 16 & 8 & 8 \\
104.83 & 1024 & 2816 & 6 & 16 & 16 \\
104.51 & 512 & 1408 & 29 & 8 & 8 \\
152.60 & 1280 & 3456 & 6 & 20 & 20 \\
153.53 & 512 & 1408 & 45 & 8 & 8 \\
\hline
\end{tabular}
\caption{Phi-3-mini draft configurations for latency benchmarks. \#Params values exclude the embeddings table parameters}
\label{tab:latency-draft-arch-config}
\end{table*}

\begin{table*}[h]
\centering
\begin{tabular}{@{}cccccc@{}}
\hline
Name & Hidden size & Intermediate size & \# Layers & \#Attention heads & \# Key-value heads\\ 
\hline
Phi3-draft-256&	256	&768	&32	&4&	4\\
Phi3-draft-384&	384	&1024	&12	&6&	6\\
Phi3-draft-448&	448	&1280	&8	&7&	7 \\
Phi3-draft-576&	576	&1536	&4	&9	&9\\
Phi3-draft-640&	640	&1792	&3	&10&	10\\
Phi3-draft-768&	768	&2048	&2	&12	&12\\
\textbf{Phi3-draft-base}& \textbf{512}	&\textbf{1408}	&\textbf{6}	&\textbf{8}	&\textbf{8}\\
\hline
\end{tabular}
\caption{Configurations of models sharing the parameter budget with the Phi3-mini 50M draft configuration}
\label{tab:latency-vs-acceptance-draft-arch-config}
\end{table*}

\begin{table*}[t]
\centering
\begin{tabular}{@{}llcccccc@{}}
\hline
Model & \multicolumn{2}{c}{CNN-DM} & \multicolumn{2}{c}{TS} & \multicolumn{2}{c}{Dolly}\\
 & AR & Speedup & AR & Speedup & AR & Speedup
 \\ \hline
Phi3-mini-256	&0.3319	&1.2481	&0.3182	&1.2223	&0.2681	&1.1283 \\
Phi3-mini-384	&0.3170	&1.5616	&0.3071	&1.5379	&0.2097	&1.3039 \\
Phi3-mini-448	&0.3255	&1.6222	&0.3009	&1.5616	&0.2177	&1.3568 \\ 
\textbf{Phi3-mini-512}	&0.3277	&1.6941	&0.3021	&1.6284	&0.2351	&1.4568 \\
Phi3-mini-576	&0.3230	&1.7066	&0.2905	&1.6220	&0.2222	&1.4444 \\
Phi3-mini-640	&0.3104	&1.6918	&0.2836	&1.6212	&0.2251	&1.4676 \\
Phi3-mini-768	&0.2236	&1.4756	&0.2713	&1.6020	&0.2006	&1.4149 \\
 \hline
\end{tabular}%
\caption{AR and Estimated speedup results for models obtained by modifying depth and width to stay within the same parameter budget as {Phi3-mini-base} draft model }
\label{tab:ar-vs-speedup-const-budget}
\end{table*}

\begin{figure*}[t]
    \centering
    \includegraphics[width=\textwidth]{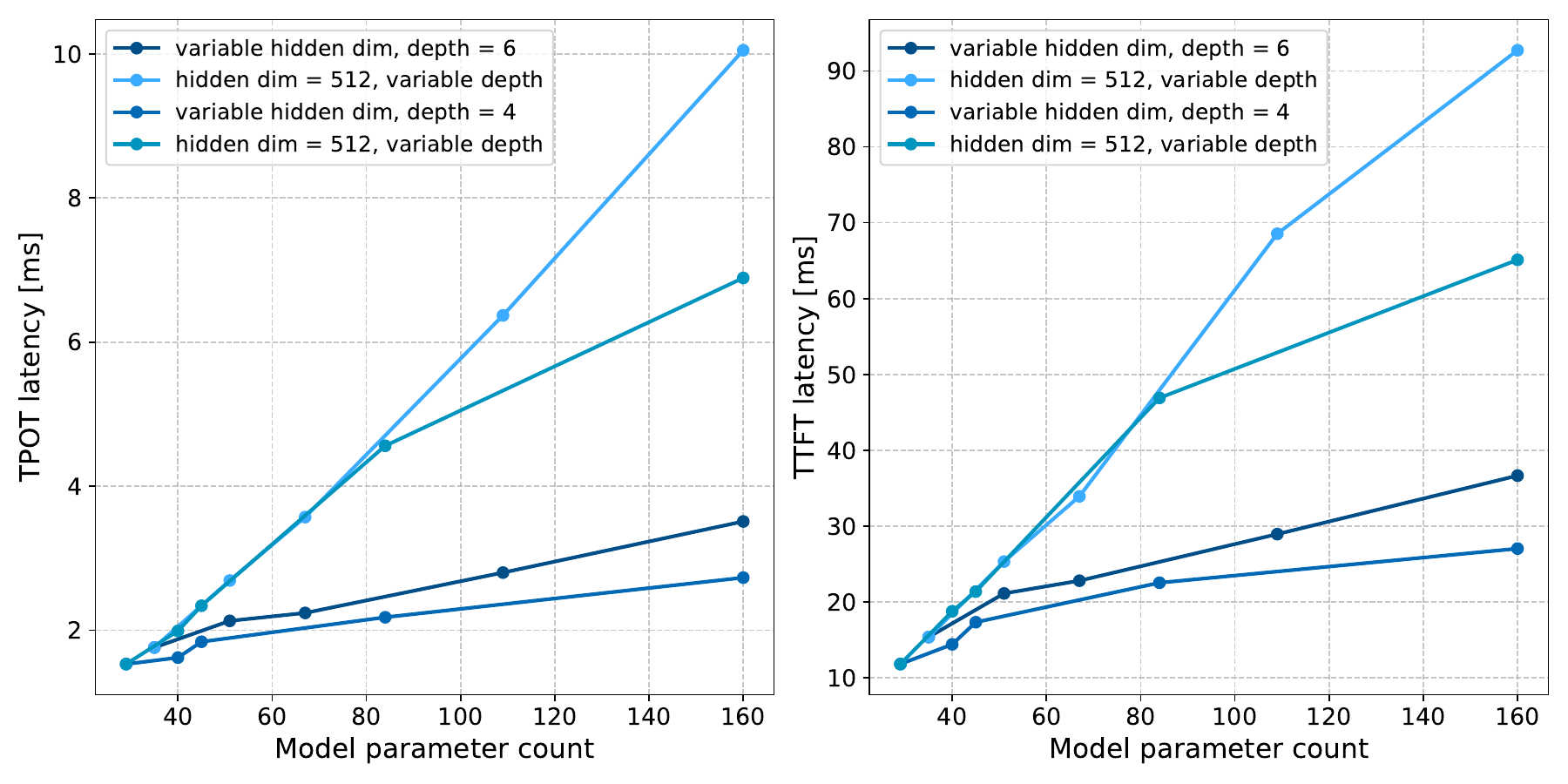}
    \caption{First and second token latencies of models obtained by varying either a hidden dimension or a number of layers}
    \label{fig:latency-layers-vs-hidden-complete}
\end{figure*}

\section{FastDraft additional results}
\label{app:full-pipeline}
Table~\ref{tab:full-pipeline-multinomial} summarizes results of each stage of the FastDraft scheme using greedy sampling.

\begin{table*}[t]
\centering
\resizebox{\textwidth}{!}{%
\begin{tabular}{@{}llcccccccc@{}}
\hline
\multirow{2}{*}{Model} & \multirow{2}{*}{Type} & \multicolumn{2}{c}{CNN-DM} & \multicolumn{2}{c}{TS} & \multicolumn{2}{c}{Dolly} & \multicolumn{2}{c}{HumanEval}\\
 &  & $\gamma=3$ & $\gamma=5$ & $\gamma=3$ & $\gamma=5$ & $\gamma=3$ & $\gamma=5$ & $\gamma=3$ & $\gamma=5$ \\ \hline
\multirow{3}{*}{Phi3-mini 50M} & \( PT\) & 0.350 & 0.246 & 0.295 & 0.196 & 0.258 & 0.174 & 0.312 & 0.221\\
 & \( PT \rightarrow CP \) & 0.349 & 0.244 & 0.304 & 0.204 & 0.244 & 0.163 & \textbf{0.672} & \textbf{0.563}\\
 & \( PT \rightarrow CP \rightarrow FT\) & \textbf{0.399} & \textbf{0.289} & \textbf{0.321} & \textbf{0.217} & \textbf{0.390} & \textbf{0.279} & 0.663 & 0.553\\ \hline
\multirow{3}{*}{Llama3.1 150M} & \( PT\) & 0.298 & 0.204 & 0.243 & 0.162 & 0.257 & 0.171 & 0.282 & 0.198\\
 & \( PT \rightarrow CP \) & 0.300 & 0.203 & 0.250 & 0.166 & 0.284 & 0.193 & 0.658 & 0.546\\
 & \( PT \rightarrow CP \rightarrow FT\) & \textbf{0.327} & \textbf{0.228} & \textbf{0.271} & \textbf{0.181} & \textbf{0.350} & \textbf{0.247} & \textbf{0.700} & \textbf{0.593}\\
 \hline
\end{tabular}%
}
\caption{AR results for FastDraft stages, performed in subsequent order: Pre-training (PT), Continued Pre-training (CP) and Fine-tuning (FT) using greedy decoding}
\label{tab:full-pipeline-greedy}
\end{table*}

\end{document}